%% file: main.tex
\documentclass{article} 
\usepackage{iclr2020_conference,times}

\input{math_commands.tex}

\usepackage[utf8]{inputenc} 
\usepackage[T1]{fontenc}    
\usepackage{hyperref}       
\usepackage{url}            
\usepackage{booktabs}       
\usepackage{amsfonts}       
\usepackage{nicefrac}       
\usepackage{microtype}      
\usepackage{xcolor}
\usepackage{sidecap}
\sidecaptionvpos{figure}{c}

\usepackage{graphicx}
\usepackage{bm}
\usepackage{amsmath}
\usepackage{multirow}
\usepackage{subcaption}
\usepackage{pbox}
\usepackage{float}
\usepackage{arydshln}
\usepackage{calc}
\usepackage{bbm}

\title{Stochastic Prototype Embeddings}

\author{Tyler R.~Scott \\
University of Colorado, Boulder \& Sensory Inc. \\
\texttt{tysc7237@colorado.edu} \\
\And
Karl Ridgeway \\ 
University of Colorado, Boulder \\
\texttt{karl.ridgeway@colorado.edu} \\
\And
Michael C.~Mozer \\
Google Research \& University of Colorado, Boulder \\
\texttt{mcmozer@google.com}
}

\iclrfinalcopy 
\begin{document}

\maketitle

\begin{abstract}
Supervised deep-embedding methods project inputs of a domain to a representational space in which same-class instances lie near one another and different-class instances lie far apart. We propose a probabilistic method that treats embeddings as random variables. Extending a state-of-the-art deterministic method, Prototypical Networks \citep{Snell2017}, our approach supposes the existence of a class prototype around which class instances are Gaussian distributed.  The prototype posterior is a product distribution over labeled instances, and query instances are classified by marginalizing relative prototype proximity over embedding uncertainty.  We describe an efficient sampler for approximate inference that allows us to train the model at roughly the same space and time cost as its deterministic sibling.  Incorporating uncertainty improves performance on few-shot learning and gracefully handles label noise and out-of-distribution inputs. 
Compared to the state-of-the-art stochastic method, Hedged Instance Embeddings \citep{Oh2019}, we achieve superior large- and open-set classification accuracy.
Our method also aligns class-discriminating features with the axes of the embedding space, yielding an interpretable, 
disentangled representation.
\end{abstract}

\section{Introduction}
\label{intro}

Supervised deep-embedding methods map instances from an input space to a latent embedding space in which same-label pairs are near and different-label pairs are far. The embedding thus captures semantic relationships without discarding inter-class structure.  In contrast, consider a standard neural network classifier with a softmax output layer trained with a cross-entropy loss. Although its penultimate layer might be treated as an embedding, the classifier's training objective attempts to orthogonalize all classes and thereby eliminate any information about 
inter-class structure.

Nearly all methods previously proposed for deep embeddings are deterministic: an instance 
projects to a single point in the embedding space. Deterministic embeddings fail to 
capture uncertainty due either to out-of-distribution inputs (e.g., data corruption) or 
label ambiguity (e.g., overlapping classes). Representing uncertainty is important for 
many reasons, including robust classification and decision making, informing downstream models, interpreting representations, and detecting out-of-distribution samples. In this article, we propose a method for discovering \emph{stochastic} embeddings, 
where each embedded instance is a random variable whose distribution reflects the uncertainty in 
the embedding space.

Our proposed method, the \emph{Stochastic Prototype Embedding (SPE)}, is an extension of the \emph{Prototypical Network (PN)} \citep{Snell2017}. 
As in the PN, our SPE assumes each class can be characterized by a prototype in the embedding space and an instance is classified based on its proximity to a prototype. In the case of the SPE, the embeddings and prototypes are Gaussian random variables, each class instance is assumed to be a Gaussian perturbation of the prototype, and a query instance is classified by marginalizing out over the embedding uncertainty. Using a synthetic data set, we demonstrate that the embedding uncertainty is related to 
both input and label noise. On a few-shot learning task, we show that the SPE significantly outperforms its state-of-the-art deterministic sibling, the PN. And on a challenging classification task, we find that the SPE outperforms \emph{Hedged Instance Embeddings (HIB)} \citep{Oh2019}, the state-of-the-art stochastic embedding method.

\section{Related Work}
\label{related_work}

Supervised embedding methods are popular in the few-shot learning literature \citep{Koch2015,Vinyals2016,Snell2017,Trian2017,Finn2017,Edwards2017,Scott2018,RidgewayMozer2018,Nikhil2018} where the goal is to classify query instances based on one or a small number of labeled exemplars of novel classes. These methods operate by embedding the queries and exemplars using a pre-trained network, and classifying each query according to its proximity to the exemplars. Embedding methods are also critical in open-set recognition domains such as face recognition and person re-identification \citep{Chopra2005,Li2014,Yi2014,Zheng2015,Schroff2015,liu2015targeting,Ustinova2016,Song2016,Wang2017}.
    
Loss functions used to obtain embeddings can be characterized according to the number of instances required to specify a loss. To describe these losses, we will use the notation $\vz_\alpha$ for an embedding of class $\alpha$. \emph{Pairwise} losses attempt to minimize within-class distances, $||\vz_\alpha - \vz'_{\alpha}||$, and maximize between-class distances, $||\vz_\alpha-\vz_\beta||$ \citep{Chopra2005,Hadsell2006,Yi2014}. \emph{Triplet} losses attempt to ensure within-class instances are closer than between-class instances, $||\vz_\alpha - \vz'_{\alpha}||< ||\vz_\alpha - \vz_\beta||$ \citep{Schroff2015,Song2016,Wang2017}. \emph{Quadruplet} losses attempt to ensure every within-class pair is closer than every between-class pair, $||\vz_\alpha - \vz'_{\alpha}||< ||\vz''_{\alpha}-\vz_\beta||$ \citep{Ustinova2016}. Finally, 
\emph{cluster-based losses} attempt to use all instances of a class 
\citep{Rippel2016,Fort2017,Song2017,Snell2017,RidgewayMozer2018}. In particular, the Prototypical Network \citep{Snell2017} computes the mean of a set of instances of a class, $\bar{\vz}_\alpha$, and ensures that additional instances of that class, $\vz_\alpha$, satisfy a proximity constraint such as $||\vz_\alpha - \bar{\vz}_\alpha|| < ||\vz_\alpha - \bar{\vz}_\beta||$. Cluster-based methods represent state-of-the-art over, in particular, pairwise and triplet losses, as one might expect given the chronology of publications.

Recently, probabilistic embedding methods have begun to appear. 
\citet{Allen2019} extend PNs via Bayesian nonparametric methods that treat each prototype as a mixture distribution, though they do not 
explore uncertainty in the embedding space nor leverage the embedding 
to handle noisy inputs and noisy labels, which is a significant aspect 
of our work. 
\citet{Vilnis2015} propose an \emph{unsupervised} method for learning density-based word embeddings, where each embedding is represented by a Gaussian distribution; however this work is not comparable to our
\emph{supervised} method.
Deep Variational Transfer \citep{Belhaj2018} is a generative form of the  discriminative model we propose; this work has the drawback that it needs to model the input distribution. Authors of this work used their approach for covariate shift, a somewhat different problem than we tackle. 

Two prior methods have been proposed for discovering stochastic embeddings in a
supervised setting, i.e., for few-shot and open-set recognition. The
\emph{Hedged Instance Embedding (HIB)} \citep{Oh2019} 
utilizes a probabilistic alternative to the contrastive loss and is trained
using a variational approximation to the information bottleneck principle. 
HIB is critically dependent on a constant, $\beta$, that determines 
characteristics of an information bottleneck (i.e., how much of the input 
entropy is retained in the embedding).  Choosing this constant is a matter of art.
The \emph{Oracle-Prioritized Belief Network (OPBN)} \citep{Karaletsos2016}
is a generative model that learns a joint distribution over inputs and 
oracle-provided triplet constraints.  The OPBN was not tested on
few-shot and open-set recognition because it requires extensions to be
applied to classification tasks.  In the deterministic setting, \citet{Scott2018}
argue that cluster-based methods outperform pairwise and triplet methods; thus,
we have reason to expect that in a stochastic setting, a cluster-based method like
the one we propose in this article, SPE, will outperform pairwise (HIB) and 
triplet (OPBN) methods.


\section{The Model}
\label{method}

The SPE assumes that the latent representation, $\vz$, is a Gaussian RV conditioned on the input, $\vx$:
\begin{equation}
\label{eq:pzx}
p(\vz|\vx) =\mathcal{N}(\vz; \vmu_x, \vsigma^2_x \mI)
\end{equation}
with mean, $\vmu_x$, and variance, $\vsigma^2_x$, computed by a deep neural network, similar to a Variational Autoencoder \citep{Kingma2014}. 
The classification, $y$, in turn is conditioned on $\vz$, with $p(y|\vz)$ taking the same form as in the original PN \citep{Snell2017}, to be described shortly. Given an input, a class prediction is made by marginalizing over the embedding uncertainty:
\begin{equation}
\label{eq:pyx}
p(y \vert \vx) = \int_\vz p(y \vert \vz) p(\vz \vert \vx) d \vz ,
\end{equation}

\begin{figure}[t!]
    \centering
    \includegraphics[width=\linewidth]{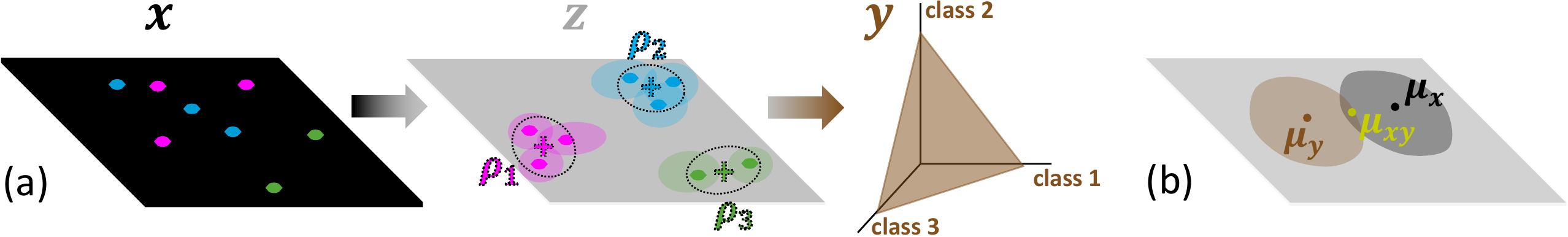}
    \caption{ (a) Illustration of the stochastic prototype embedding. The model learns a mapping 
    from input space, $\vx$, to embedding space, $\vz$, in which same-class instances are 
    near and different-class instances are far. Embeddings are represented as Gaussian 
    random variables.
    Prototypes, noted as $+$ symbols in the embedding, are formed via a 
    confidence-weighted average of the embeddings of instances known to belong
    to a class (support instances). Prototype uncertainty is depicted with the dotted ovals.
    Given the prototypes, a prediction of class $y$ is made for a query instance by
    marginalizing a softmax prediction over the embedding space. (b) Depiction of intersection sampler.}
    \label{fig:spe_model}
\end{figure}

Figure~\ref{fig:spe_model}a depicts the relationship between the input, latent, and class representations.
We train the SPE using the standard few-shot learning paradigm, consisting of a sequence of \emph{episodes}, each with $m$ instances of $n$ classes. We split the $m\times n$ instances into $k \times n$ \emph{support} examples, defining a set $S$, and $(m-k) \times n$ \emph{query} examples. The support instances for each class $c$, $S_c \in S$, are used to determine the class prototype, $\vrho_c$, and the query instances are evaluated to predict class label (Equation~\ref{eq:pyx}).

\subsection{Forming class prototypes}
\label{forming_prototypes}

In the SPE, each class $y$ has an associated prototype, $\vrho_y$, in the embedding space, and each instance $i$ of class $y$, denoted $\vx_i$, projects to an embedding, $\vz_i$, in the neighborhood of $\vrho_y$ such that:
\begin{equation}
\label{eq:epsilonnoise}
\vrho_y = \vz_i + \vepsilon,
\text{ where } \vepsilon \sim \mathcal{N}(\bm{0}, \sigma^2_{\epsilon}\mI).
\end{equation}
We assume that the prototype is consistent with all support instances, allowing us to express the likelihood of $\vrho_y$ as a product distribution:
\begin{equation}
\label{eq:product_distribution}
p(\vrho_y \vert S_y) = \frac{\prod_{i \in S_y} p(\vrho_y \vert \vx_i)}{\int_\vrho \prod_{i \in S_y} p(\vrho \vert \vx_i) d\vrho}.
\end{equation}
Because $p(\vrho_y|\vx_i)$ is Gaussian, the resulting product is too:
\begin{equation} \textstyle
\vrho_y | S_y \sim \mathcal{N} (\vmu_y, \vsigma^2_y \mI) \text{ ~with~  }
\vsigma^2_y = \left( \sum_{i\in S_y} \hat{\vsigma}^{-2}_{x_i} \right)^{-1} 
\text{ and~~ } 
\vmu_y = \vsigma^2_y \circ \left( \sum_{i\in S_y}
		  \hat{\vsigma}^{-2}_{x_i} \circ \vmu_{x_i} \right) ,
\label{eq:confidence_weighted_prototype}
\end{equation}
where 
$\hat{\vsigma}^2_{x_i} = \vsigma^2_{x_i} + \sigma^2_\epsilon$
and
$\circ$ denotes the Hadamard product.
Essentially, the prototype is a confidence-weighted average of the support instances.
This formulation has a clear advantage over the deterministic PN, which is premised on an unweighted
average, because it de-emphasizes noisy support instances.

\subsection{Prediction and approximate inference}
\label{marginalization}

We assume a softmax prediction for a query embedding, $\vz$:
\begin{equation}
\label{eq:pyz}
p(y|\vz, S) \propto 
\mathcal{N}(\vz;\vmu_y,\hat{\vsigma}^2_y \mI)
\end{equation}
with $\hat{\vsigma}^2_y = \vsigma^2_y + \sigma^2_\epsilon$ as before, yielding the class posterior for query $\vx$:
\begin{equation}
\begin{split}
\label{eq:classif}
p(y|\vx, S) = 
\int_\vz \mathcal{N}(\vz;\vmu_{x},\vsigma^2_{x}\mI ) ~\frac{\mathcal{N}(\vz;\vmu_y,\hat{\vsigma}^2_y\mI )}{\sum_c \mathcal{N}(\vz;\vmu_c,\hat{\vsigma}^2_c\mI )} ~d \vz .
\end{split}
\end{equation}

The class distribution is equivalent to that produced by the deterministic PN as $\vsigma_{x}^2 \rightarrow 
\boldsymbol{0}$ when $\vsigma_{y}^2 = \vsigma_{y'}^2$ for all class pairs $(y, y')$.
However, in the general case, the integral has no closed form solution; thus,
we must sample to approximate $p(y|\vx,S)$, both for
training and evaluation. We employ two samplers,
which we refer to as \emph{na\"ive} and \emph{intersection}.

\subsubsection{Na\"ive sampling}
A direct approach to approximating the class posterior is to express Equation \ref{eq:pyx} as an expectation, $\mathbb{E}_{\vz \sim p(\vz|\vx)} \left[p(y|\vz,S) \right]$, and to replace the expectation with the average over a set of samples. We utilize the reparameterization trick of \citet{Kingma2014} to train the model. Although this is the simplest approach, it is sample-inefficient during training, and when the number of samples is reduced, model performance is impacted.

\subsubsection{Intersection sampling}

In Equation \ref{eq:classif}, the product of Gaussian densities in the numerator can be rewritten:
\begin{equation}
\mathcal{N}\left(\vz;\vmu_{x},\vsigma^2_{x}\mI\right)~ \mathcal{N}\left(\vz;\vmu_y,\hat{\vsigma}^2_y \mI\right) = \mathcal{N}\left(\vz;\vmu_{xy},\vsigma^2_{xy}\mI\right)~ \mathcal{N}\left(\vmu_x;\vmu_y,(\vsigma^2_x  + \hat{\vsigma}^2_y)\mI\right) ,
\label{eq:subst}
\end{equation}
where 
$\vsigma^2_{xy}=(\vsigma^{-2}_x + \hat{\vsigma}^{-2}_y)^{-1}$ and~ 
$\vmu_{xy} = \vsigma^2_{xy} \circ (\vsigma^{-2}_x \circ \vmu_x + \hat{\vsigma}^{-2}_y \circ \vmu_y).$
Substituting Equation \ref{eq:subst} into Equation \ref{eq:classif}, 
\begin{equation}
\label{eq:intersec_classif}
p(y|\vx, S) =~ \mathcal{N} \left(\vmu_{x};\vmu_y,  (\vsigma^2_x + \hat{\vsigma}^2_y )\mI \right) \mathbb{E}_{\vz \sim \mathcal{N}\left(\vmu_{xy},\vsigma^2_{xy} \mI\right)}
\left[\sum_c \mathcal{N}(\vz;\vmu_c,\hat{\vsigma}^2_c\mI ) \right]^{-1}.
\end{equation}
By approximating the expectation with samples from 
$\mathcal{N}\left(\vmu_{xy},\vsigma^2_{xy} \mI\right)$, we obtain a sampler that focuses on the intersection of the input distribution and a given class distribution, as illustrated in Figure~\ref{fig:spe_model}b. During training with a cross-entropy loss, we need only sample for the known (target) class $y$. As we will demonstrate, this method is more robust and significantly more
sample efficient than the na\"ive sampler, requiring only a \emph{single} sample to train effectively. 

\section{Experimental Results}
\label{results}
We report on three sets of experiments.
In Section~\ref{results:synth}, we demonstrate, using a synthetic data set, that
SPE infers the generative structure of a domain, disentangles class-discriminating features, and provides meaningful estimates of label uncertainty and input noise.
In Section~\ref{results:omniglot}, we show that SPE obtains state-of-the-art results on few-shot learning via a comparison
to its deterministic sibling, PN, the previous state-of-the-art method. We evaluate on a standard
data set used to compare methods in the few-shot learning literature, Omniglot \citep{omniglot}.
In Section~\ref{results:multimnist}, we show that SPE obtains state-of-the-art results on large-set classification via a comparison
to the only other fully developed stochastic method for supervised embeddings, HIB \citep{Oh2019}. 
We evaluate on the only data set that \citet{Oh2019} used to explore HIB, a multi-digit variant of MNIST.
For details regarding network architectures and hyperparameters, see Appendix \ref{architectures}, and for simulation details, including the choice of initialization for $\sigma_\epsilon^2$, see Appendix \ref{sim_details}.

\subsection{Synthetic color-orientation data set}
\label{results:synth}
\begin{figure}[t!]
    \centering
    \includegraphics[height=1.5in]{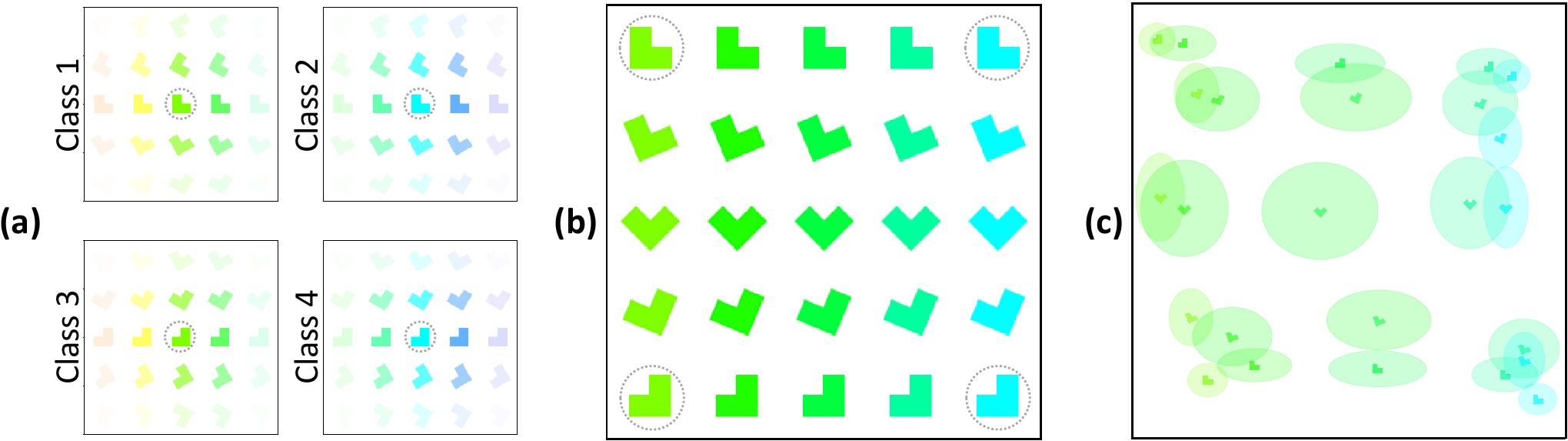}
    \caption{(a) Samples from the four classes in our synthetic data set. In each plot, class centroids are circled, along with samples 
    spanning $\pm2$ standard deviations in both orientation and color.
    A sample's transparency is set according to its class-conditional likelihood.
    Both dimensions can be coded as directional variables. The class centroids on 
    each dimension are $90^\circ$ apart with standard deviation of $30^\circ$.
    (b) A set of examples, with the four class centroids located in the corners and other examples
    obtained by  linear interpolation in the generative space. 
    (c) The $2$D stochastic prototype embedding for the examples in (b). The shape is plotted at the 
    mean of $p(\vz | \vx)$, and the outlines of the ovals represent equiprobability
    contours at $0.4$ standard deviations.}
    \label{fig:synthetic_examples_embedding}
\end{figure}

The data set consists 
of $64\times64$ pixel images of `L' shapes, with four classes that are distinguished by
orientation, color, or both (Figure~\ref{fig:synthetic_examples_embedding}a). 
Instances are sampled from a class-conditional isotropic Gaussian distribution in the generative space.
(The isotropy of these qualitatively different dimensions
comes from the fact that both can be mapped as directional quantities.)
Because classes overlap on both color and orientation dimensions, elicited embeddings  
should indicate increased uncertainty near class boundaries.
Full details of the synthetic data set can be found
in Appendix \ref{synthetic_ds}.

\begin{SCfigure}[1.][b!]
    \centering
    \includegraphics[height=1.35in]{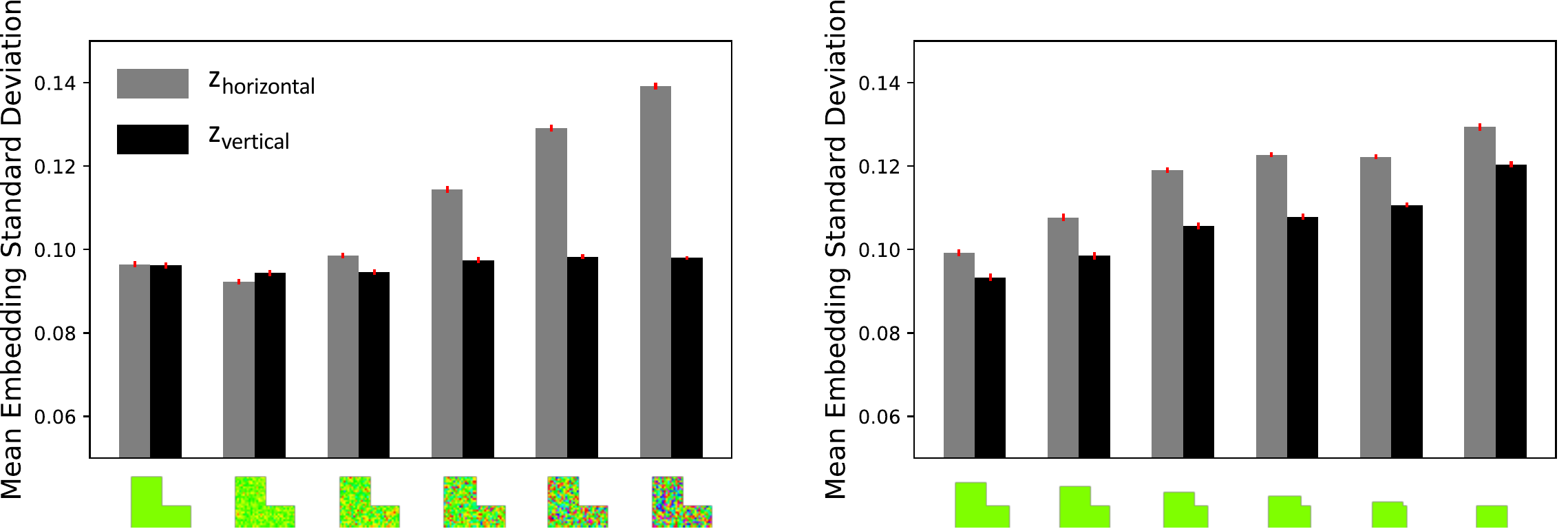}
    \hspace{.1in}
    \caption{Synthetic data set: uncertainty on the two embedding dimensions as 
    it becomes more difficult to discern the hue (left) and orientation (right).}
    \label{fig:inputnoise}
\end{SCfigure}

We trained a two-dimensional, intersection-sampling SPE on samples from this domain,
using two instances per class to form prototypes.
Classification accuracy of held-out samples is approximately $86\%$.
Accounting for class overlap, a Bayes optimal classifier has an accuracy of approximately
$87\%$.
For visualization, Figure~\ref{fig:synthetic_examples_embedding}b presents a $5\times5$ array of
examples with the class centroids in the corners and the other examples obtained by linear interpolation in the generative space. The resulting embeddings are presented in 
Figure~\ref{fig:synthetic_examples_embedding}c. Although the correspondence between Figures~\ref{fig:synthetic_examples_embedding}b and ~\ref{fig:synthetic_examples_embedding}c
seems trivial (mirror one set along the horizontal axis to obtain the other set), remember that
the input space is $64\times64$ dimensional and the latent space is $2$ dimensional.
The network has captured the structure of the domain by disentangling the two factors of variation.
Further, the embedding variance encodes label ambiguity;
instances halfway between two classes on one dimension have maximal
variance along that dimension.
Label ambiguity is one type of uncertainty. An equally important source of uncertainty
comes from noisy or out-of-distribution (OOD) inputs. We examined OOD inputs generated in
two different ways. In the left panel of  Figure~\ref{fig:inputnoise},
we show the consequence of adding pixel hue noise to the four class centroids. Only
one of these centroids is shown along the abscissa, but all four are used to make the graph,
with many samples per noise level.
The grey and black bars in the graph indicate variance on the horizontal and vertical
dimensions of the embedding space, respectively. As pixel hue noise increases, uncertainty
in color grows but uncertainty in orientation does not. In the right panel of Figure~\ref{fig:inputnoise}, we show the consequence of shortening the leg-length of the shape. Shortening the legs removes cues that can be used both for determining color
and orientation. As a result, the uncertainty grows on both dimensions.

\subsection{Omniglot}
\label{results:omniglot}
The Omniglot data set contains images of labeled, handwritten characters 
from diverse alphabets. Omniglot is one of the standard data sets for comparing 
methods in the few-shot learning literature. The data set contains $1623$ unique 
characters, each with $20$ instances. Following \citet{Snell2017}, 
each grayscale image is resized from $32 \times 32$ to $28 \times 28$, and
we augment the original classes with all $90^\circ$ rotations, resulting in $6492$ total classes.
We train PNs and SPEs episodically, where a training episode contains $60$ randomly sampled classes and 
$5$ query instances per class.

\begin{SCfigure}[1.][t!]
\includegraphics[scale=0.17]{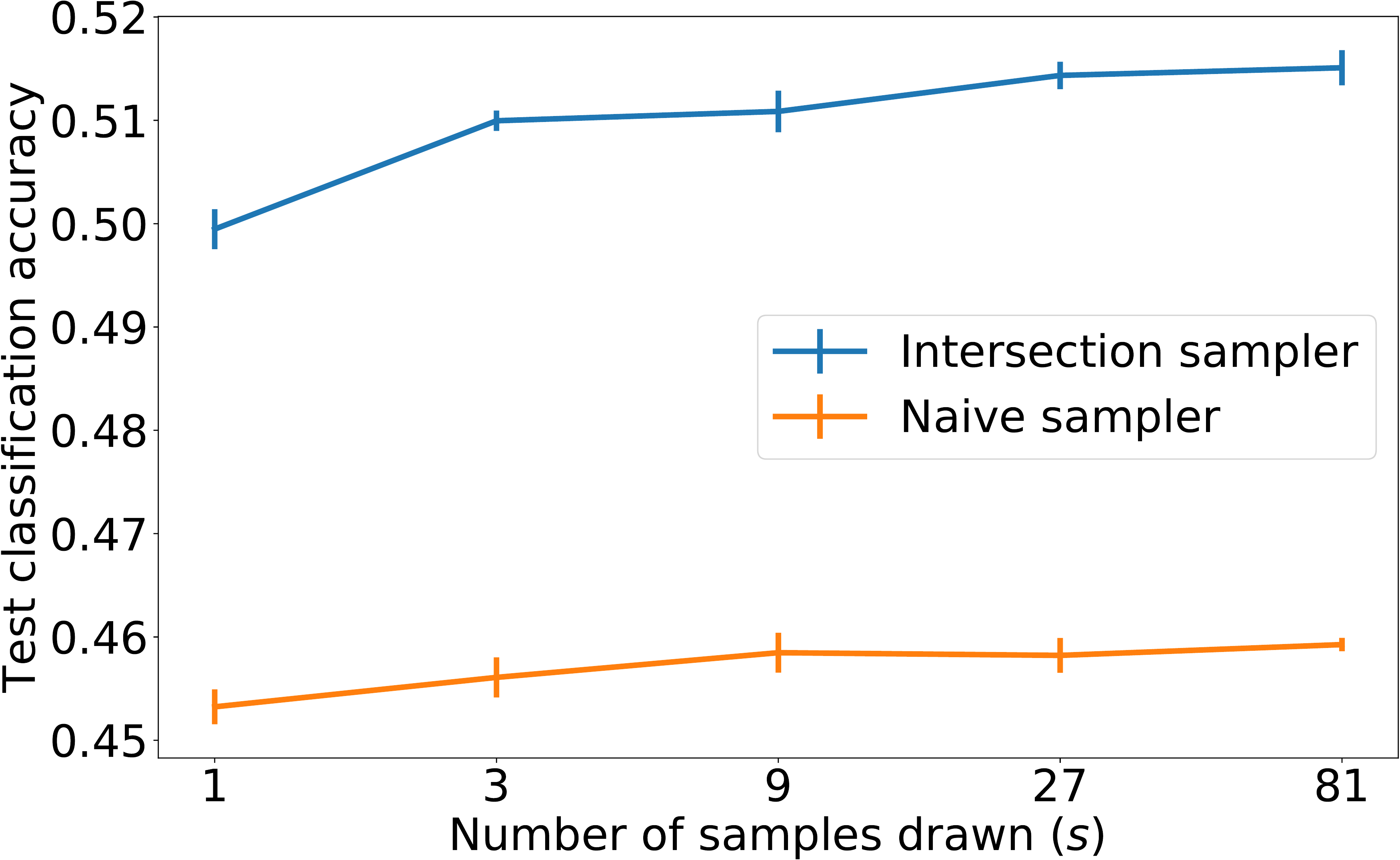}
\caption{Test classification accuracy as a function of number of training samples per query instance for a na\"ive-sampling and intersection-sampling $2$D SPE on a $1$-shot, $20$-class Omniglot task. Performance is a mean over $5$ replications of running the model, showing $\pm1$ standard error of the mean.}
\label{fig:naive_vs_intersection}
\end{SCfigure}

\begin{SCfigure}[1.][b!]
\centering
\includegraphics[scale=0.3]{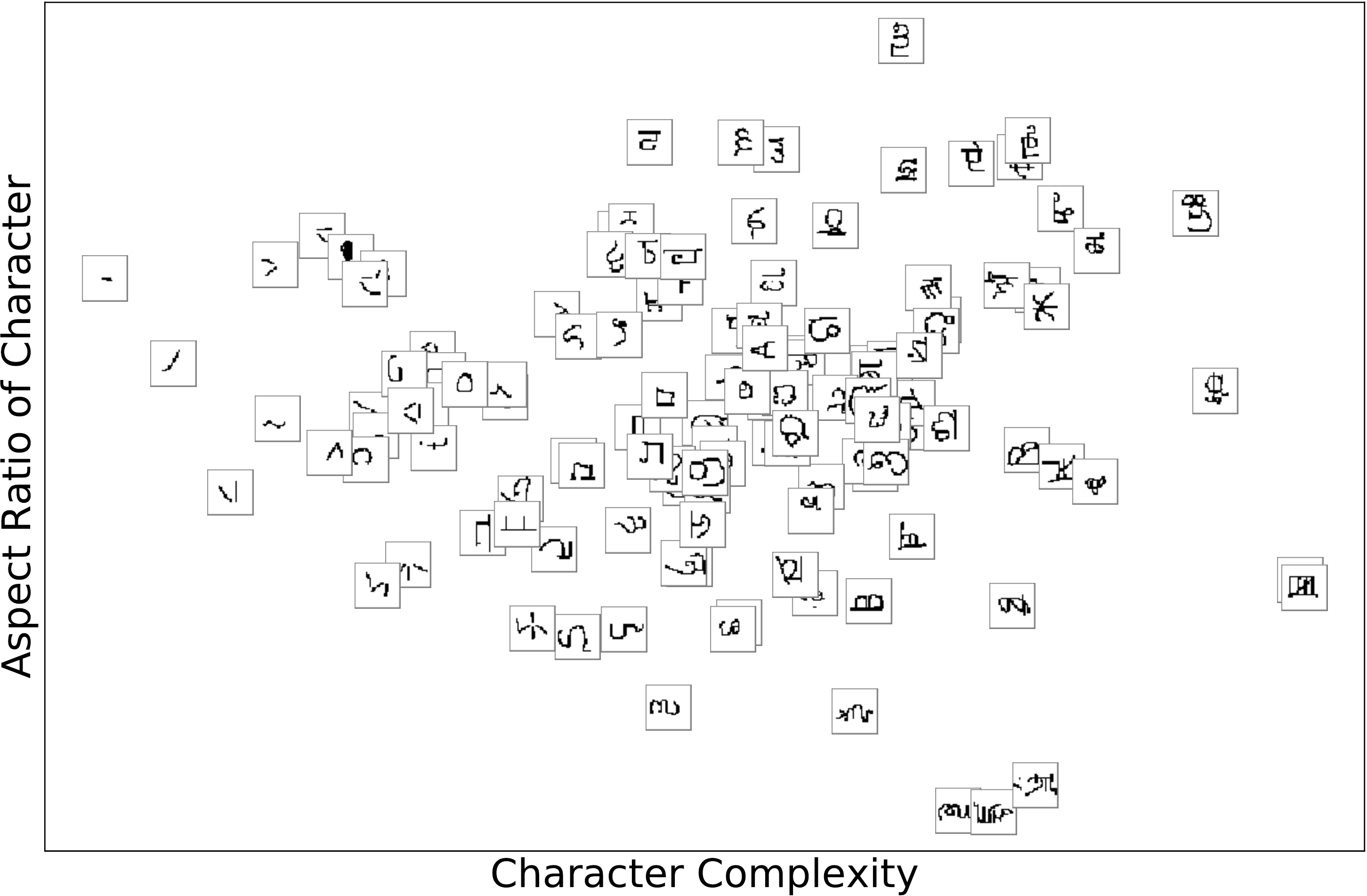}
\caption{Two-dimensional embedding learned by the SPE on the Omniglot test set. Each square thumbnail image in the figure is a randomly-sampled instance from one of $125$ randomly-sampled test classes and the location of the image represents the location of the class prototype. The images have a gray bounding box for visualization purposes only.}
\label{fig:omniglot_embedding}
\end{SCfigure}

To compare the relative effectiveness of na\"ive and intersection samplers, we train the SPE on Omniglot
varying both the sampler and the number of samples drawn per training 
query, denoted by $s$. We evaluate in a $1$-\emph{shot} $20$-\emph{class} setting, where 
shot refers to the number of support examples used to compute each prototype.
Figure  \ref{fig:naive_vs_intersection} 
shows test  classification accuracy as the number of samples drawn per 
training trial ($s$) increases.
As we previously stated, the intersection-sampling SPE is far more sample efficient, to the point that
the intersection sampler with $s=1$ outperforms the na\"ive sampler with $s=81$.
We have verified that the pattern in Figure \ref{fig:naive_vs_intersection} is 
consistent across simulations; consequently, we present only intersection-sampling SPE results 
in the remainder of the article, and all SPEs are
trained with a single sample ($s=1$) per query. This choice causes the SPE to be on par with 
the PN in time and space requirements, even though using more samples may boost
classification accuracy, as suggested by the trend in Figure~\ref{fig:naive_vs_intersection}.

Figure \ref{fig:omniglot_embedding} is a visualization of a $2$D embedding learned by the intersection-sampling SPE on Omniglot. All classes shown in the figure were held-out during training. 
Omniglot characters clearly vary along more than two dimensions, so a $2$D SPE cannot 
learn a fully-disentangled representation as it did with the synthetic data set. 
However, we can still interpret the axes of the embedding. 
The horizontal axis appears to represent character complexity, with single-stroke characters on the left 
and many-stroke characters on the right.
The vertical axis appears to encode the aspect ratio of the characters, with horizontally extended
characters on the bottom and vertically extended characters on the top.

Figure \ref{fig:combined_bar_results}a compares the PN and SPE with $2$D embeddings
on Omniglot test classes. Each bar is the mean accuracy across four conditions: $1$-shot/$5$-class,
$5$-shot/$5$-class, $1$-shot/$20$-class, and $5$-shot/$20$-class. The first pair of bars perform
the standard comparison in which the (1 or 5 instance) support set is used to obtain an embedding for
each class, prototypes are formed, and query instances are classified. SPE is reliably better than the PN. 
Because the Omniglot data are carefully curated, the instances have little noise and
therefore offer  little opportunity to leverage SPE's assessment of uncertainty. 
Consequently, we corrupted instances by masking out
rectangular regions of the input, as proposed by \citet{Oh2019}. 
(See Appendix~\ref{corruption_procedure} for details.) The second and
third sets of bars in Figure~\ref{fig:combined_bar_results}a correspond to the situations where the
support and query instances are corrupted, respectively. SPE's advantage over PN increases significantly
when the support instances are corrupted due to the fact that SPE's confidence-weighted prototypes
(Equation~\ref{eq:confidence_weighted_prototype})
discount noisier support examples. Although the SPE is still superior when only
the query is corrupted, the benefit is small. We also compared PN and SPE using a $64$D embedding, 
but with high dimensional embeddings, both methods are near ceiling on this data set, resulting in 
comparable performance between the two methods. (See Appendix~\ref{appendix:tabular_results} for additional
results, broken down by condition.)

To emphasize, SPE outperforms the PN, arguably the leading few-shot learning method, especially when
inputs are corrupted, at essentially the same computational cost for training.
And by providing an estimate of uncertainty associated with embedded instances, the SPE
offers the possibility of detecting OOD samples and informing downstream systems that
operate on the embedding. 

\subsection{N-digit MNIST}


\label{results:multimnist}
The $N$-digit MNIST data set was proposed to evaluate HIB \citep{Oh2019}; it is 
formed by horizontal concatenation of $N$ MNIST digit images. The resulting 
images are  $28 \times 28N$. 
To compare with HIB, we study $2$- and $3$-digit MNIST, and
use a network architecture identical to that in \cite{Oh2019}.
\citet{Oh2019} split the data into a training set (with $70\%$ of the 
total classes), a \emph{seen} test set, and an \emph{unseen} test set. 
For $2$-digit MNIST, the seen test set has the same $70$ of $100$ classes 
as the training set and the unseen test set has the remaining $30$ classes. 
For $3$-digit MNIST, the training set has $700$ classes, the seen and unseen
test sets each have a sample of $100$ of the $700$ seen or $300$ unseen classes,
respectively. We use the same train and test data splits as \cite{Oh2019}, but 
we further divide the training split to include a validation set for early 
stopping.

\begin{figure}[t!]
    \centering
    \includegraphics[scale=0.175]{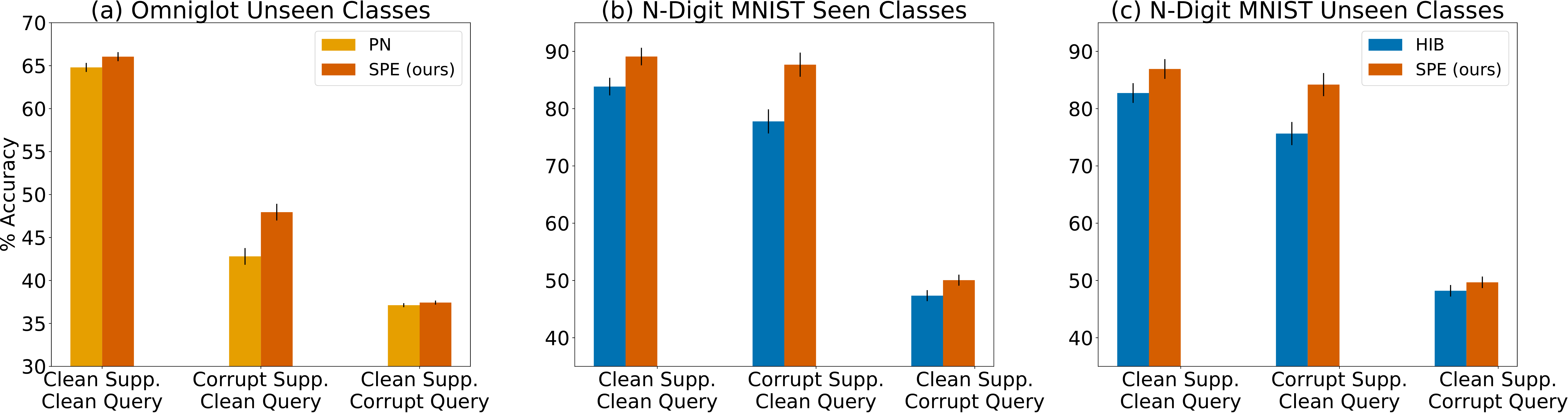}
    \caption{(a) Comparison of few-shot accuracy on Omniglot test classes for the PN 
    \citep{Snell2017} and our  SPE.
    (b) Comparison of test accuracy on seen classes for $2$ and $3$-digit MNIST for HIB \citep{Oh2019} 
    and our SPE.  (c) Same as (b) except for unseen classes. In (a)-(c),
    error bars reflect $\pm 1$ standard error of the mean, corrected to  remove cross-condition 
    variance \citep{MassonLoftus2003}.} 
    \label{fig:combined_bar_results}
\end{figure}

\begin{figure}[t!]
\centering
\includegraphics[scale=0.18]{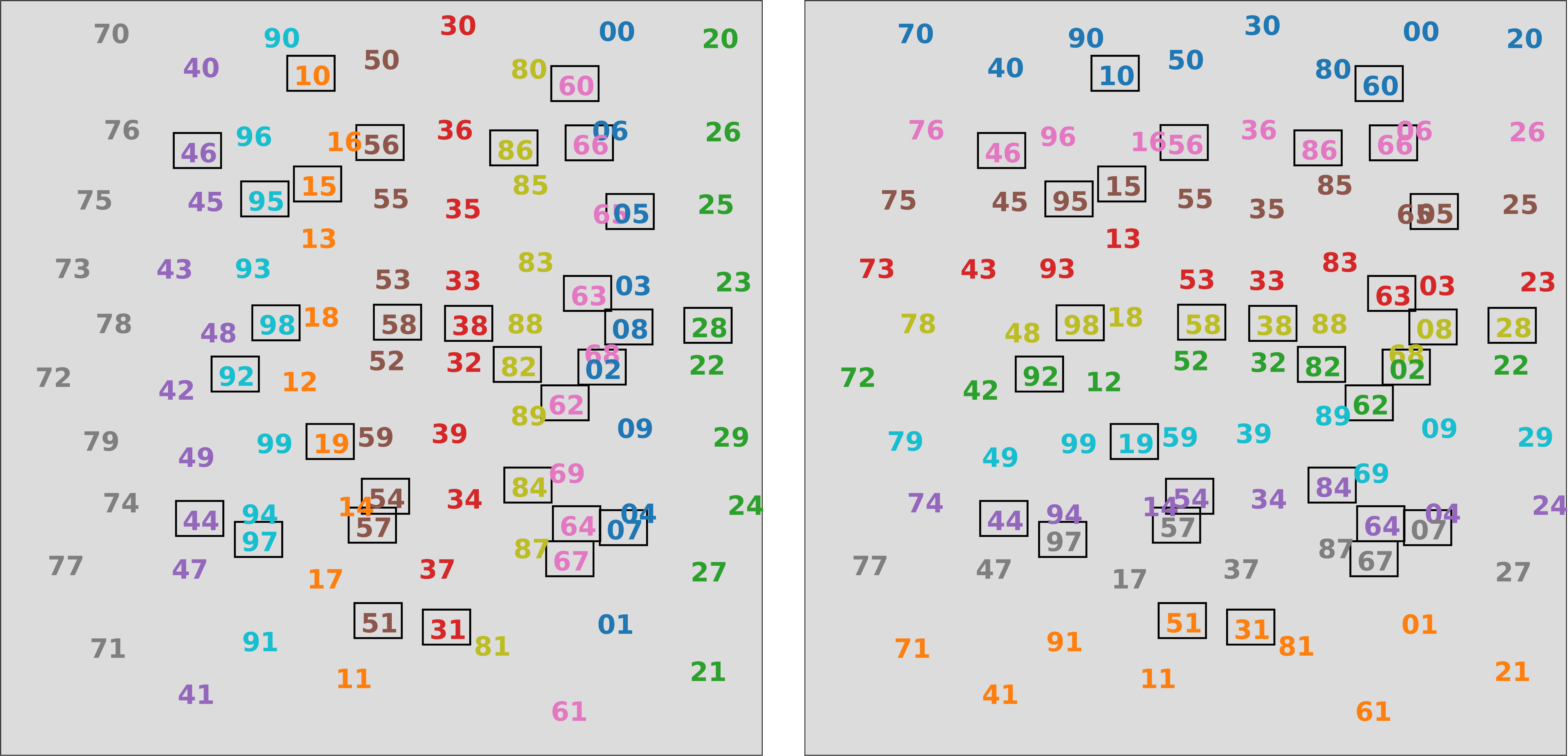}
\caption{Two-dimensional embedding learned by the SPE on the $2$-digit MNIST test set. 
A class is specified by a two-digit number.
In both figures, the location of the class corresponds to the mean of the prototype in the test set using $140$ support instances. The digits surrounded by a black border are classes that were not seen during training. 
In the left and right figures, 
the prototypes are colored according to the first and second digit of the class, respectively.}
\label{fig:2digit_mnist_embedding}
\vskip -0.1in
\end{figure}

Figure \ref{fig:2digit_mnist_embedding} shows two views of the $2$D embedding learned by the SPE on the $2$-digit MNIST test set. Each number is a class label; for example, $71$, located in the lower left of the embedding, is the class in which the first of the two MNIST digits is a $7$ and the second is a $1$. 
The location of a label in the space corresponds to the mean of its prototype. 
In the left plot, each class is colored according to the first digit. 
The right plot is the same embedding, but each prototype is colored according 
to the second digit. 
The SPE learns an 
incredibly robust factorial representation in which the horizontal dimension represents the first digit 
of a class and the vertical dimension represents the second digit. 
A black bounding box indicates the unseen test classes, classes not presented during training.
Impressively, the unseen test classes are embedded in exactly the positions where they
belong, indicating that the SPE can discover relationships among classes that allow it to generalize
to classes it has never seen during training.
Furthermore, the embedding has captured inter-class similarity structure by placing visually
similar digits close to one another. For example, on both the vertical and horizontal
bands, nines (teal) and fours (purple) are adjacent, and fives (brown) and threes 
(red) are adjacent.  The adjacency relationships vary a bit from one dimension of the mapping
to the other; for example, sixes (pink) are adjacent to eights (yellow) and zeros (blue) in 
the vertical bands, but adjacent to fives (brown) and zeros in the horizontal bands.
HIB is able to discover a similar structure along one dimension \citep{Oh2019}, but
the second dimension is somewhat more entangled, suggesting that 
the SPE learns a more robust representation. Additionally, embeddings for the unseen
class are not presented for HIB. The ability to sensibly embed novel classes is 
essential for any model that will be used for open-set recognition or 
few-shot learning.

Figure~\ref{fig:combined_bar_results}b,c compare $N$-digit MNIST test accuracy on seen and unseen
classes, respectively.\footnote{HIB results are from \citet{Oh2019}. We thank the authors for 
providing us results on unseen classes, which were not included in their publication.}
Each bar is the mean test accuracy across the Cartesian product of conditions specified by
the number of MNIST digits in each image, $N \in \{2,3\}$, and the dimensionality of the embedding,
$D \in \{2,3\}$.  As in the Omniglot simulation, we varied whether support and query instances were
clean or corrupted.  The SPE outperforms HIB in all six comparisons. 
In the $24$ individual conditions, SPE is worse on only $7$.  
As in the Omniglot simulation, SPE shines best when support instances may be corrupted.  
(Appendix~\ref{appendix:ndigmnist} provides tabular results by condition,
not only for HIB and SPE, but also their deterministic counterparts, contrastive loss and PN. Because
the deterministic methods perform consistently worse than the stochastic methods, we omit the
deterministic methods from the figure.)

Whereas SPE is a discriminative model with a specified classification procedure, \cite{Oh2019} had the freedom to design one. They use all available data---roughly $140$ examples per class---and perform leave-one-out $5$-nearest-neighbor classification. To be consistent with our episodic test procedure, the SPE uses only $50$ support instances per class to form prototypes.
It is particularly impressive that the SPE, based on a single stored prototype and approximately $1/3$ the labeled data, can outperform a memory-based nonparametric method that is able to model arbitrary distributions in latent space.

\section{Discussion and Conclusions}
\label{conclusion}
Our Stochastic Prototype Embedding (SPE) method outperforms a state-of-the-art
deterministic method, the Prototypical Net (PN), on few-shot learning,
particularly when support instances may be corrupted. Because the SPE reduces to the PN under certain restrictions, it seems unlikely to fare worse; but because it can handle uncertainty in both the query and support set, it has
great opportunity to improve on the PN. Many extensions have been proposed
to the PN \citep[e.g.,][]{Fort2017,Allen2019}. These extensions are 
mostly compatible with ours, and thus methods may be potentially combined to attain even stronger few-shot learning performance under uncertainty.

SPE also significantly outperforms the only existing 
alternative stochastic method,  the Hedged Instance Embedding (HIB), on a the complete battery of
large-set classification tasks used to evaluate HIB.
Beyond its performance gains, SPE has no hand tuned
parameters, whereas HIB has constant $\beta$ that determines 
characteristics of an information bottleneck (i.e., how much of the input 
entropy is retained in the embedding).  
Although one could simply set $\beta=0$, 
doing so would encourage the net to perform like a softmax classifier and discard 
all information about inter-class similarity. Such similarities are essential in 
order to generalize to unseen classes (e.g., Figure~\ref{fig:2digit_mnist_embedding}).

We proposed and evaluated an intersection sampler to train the SPE, which makes the SPE as time 
and space efficient for training as the deterministic PN, and more efficient for training 
than HIB, which relies on about $8$ samples per item. (Our evaluation method for SPE 
presently involves drawing $200$ samples from the naive sampler, though this conservative decision was arbitrary and not tuned.)

An unanticipated virtue of SPE is its ability to obtain interpretable, disentangled representations
(Figures~\ref{fig:synthetic_examples_embedding}, \ref{fig:omniglot_embedding}, 
\ref{fig:2digit_mnist_embedding}).
Because uncertainty is encoded in a diagonal covariance matrix, any classification ambiguity
maps to uncertainty in the value of individual features of the embedding. Thus, class-discriminating
feature dimensions must align with the principle axes of the embedding space.
In contrast to traditional unsupervised disentangling methods, which aim to discover the
underlying generative factors of a domain, the SPE obtains a supervised analog in which
the underlying class-discriminative factors are represented explicitly. This representation
facilitates generalization to novel unseen classes and is therefore valuable for few-shot and lifelong learning paradigms.


\bibliography{main}
\bibliographystyle{apa}

\clearpage

\appendix

\section{Network Architectures and Hyperparameters}
\label{architectures}

\subsection{Omniglot}
\label{omniglot_arch}

For all Omniglot experiments, the network consisted of four convolutional blocks. The first three blocks had a convolutional layer with $64$ filters, a $3\times3$ kernel, zero-padding of length $1$, and a stride of $1$, followed by a batch normalization layer, ReLU activation, and $2\times2$ max-pooling. The fourth and final block had a convolutional layer with $2d$ filters, a $3\times3$ kernel, zero-padding of length $1$, and a stride of $1$, followed by $2\times2$ max-pooling, where $d$ represents the dimensionality of the embedding space. The flattened output of the network is a vector of length $2d$, where the first $d$ elements were considered the mean of the Gaussian distribution and the remaining $d$ elements were the diagonal covariance entries. The weights were initialized using He initialization and the biases with the following uniform distribution: $\mathcal{U}(-\frac{1}{\sqrt{\text{fan in}}}, \frac{1}{\sqrt{\text{fan in}}})$.

All Omniglot models were trained with an initial learning rate of $0.001$ which was cut in half every $50$ epochs. The models were stopped early using a patience parameter when performance on the validation set no longer increased.

\subsection{Synthetic data}
\label{synthetic_ds}

The images in the synthetic data set are $64 \times 64$ pixels
in size. For orientation, we chose class centers at $90^{\circ}$ and $180^{\circ}$, with a standard deviation of $30^{\circ}$.
For color, we manipulated the hue and kept value and saturation constant. Like orientation, hue is a circular quantity. If hue ranges from $0$ to $360$ degrees, we chose color class centers and standard deviation in the same way as orientation.
Additionally, we add noise to a minority (15\%) of the images used
to train the model. For these, we add Gaussian noise to the hue of 
each pixel inside the shape. The standard deviation of the hue noise
was chosen uniformly between $18^{\circ}$ and $54^{\circ}$.
We also added noise to the leg lengths of the L shapes. The leg
length was chosen uniformly between 10\% and 98\% of its original
length. See Figure \ref{fig:inputnoise} for some examples.

The network followed an architecture similar to the one we used
for Omniglot, except that we added two additional blocks of
convolution, batch normalization, ReLU, and max-pooling because the images
are larger. We used $2$ instances per class to form prototypes and $8$ samples per query instance during training. We used a learning rate of 0.0001 and the models were stopped early using a patience parameter when performance on the validation set no longer increased.

\subsection{N-Digit MNIST}
\label{appendix:ndigmnist}
For all $N$-digit MNIST experiments, we constructed an architecture which we 
believe to be identical to that used for HIB MNIST experiments, 
based on code provided by
the authors \citep{Oh2019}. The network consisted of two convolutional blocks followed by two fully-connected layers. The convolutional blocks each contained a convolutional layer, followed by an ReLU activation, and $2\times2$ max-pooling. The first convolutional layer had $6$ filters, a $5\times5$ kernel, zero-padding of length $2$, and a stride of $1$. The second convolutional layer was identical to the first, but had $16$ filters instead of $6$. The output of the second convolutional block was flattened, passed through a fully-connected layer with $120$ units, an ReLU activation, and a final fully-connected layer with $2d$ units, where $d$ represents the dimensionality of the embedding space. Like the Omniglot architectures, the first $d$ entries in the output vector are treated as the mean and the remaining $d$ elements as the diagonal covariance entries. The weights were initialized using a Xavier uniform initialization and biases were initialized to zero.

The PN and SPE  are trained episodically with all performance results in the
main article measured as the mean over $1000$ random test episodes. All $N$-digit MNIST models were trained with an initial learning rate of $0.001$ which was cut in half every $50$ epochs. The models were stopped early using a patience parameter when performance on the validation set no longer increased. For $2$-digit MNIST, each episode in training, validation, and seen-class testing contained all $70$ classes and $50$ support instances per class. For testing of unseen classes, each episode contained all $30$ classes. For $3$-digit MNIST, each episode contained $100$ classes and either $20$ support instances per class for training/validation or $50$ support instances per class for seen- and unseen-class testing.

\section{Simulation Details}
\label{sim_details}

For all SPE models,
\begin{equation*}
    \sigma^2_{\epsilon} = \mathrm{softplus}\left(\gamma \right),
\end{equation*}
where $\gamma$ is a trainable parameter. We initialize $\gamma$ using the following prescription:
\begin{equation*}
    \gamma = \vert S \vert \gamma_0^{2 / d},
\end{equation*}
where $\vert S \vert$ is the number of support examples per episode during training and $d$ is the dimensionality of the embedding. We chose this prescription for two reasons: (1) as the number of support examples increases, the variance of the prototype distribution approaches zero, so scaling linearly by $\vert S \vert$ tends to provide a stronger training signal early on, and (2) the amount of noise in the projection of an embedding should scale with the dimensionality of the embedding space as to maintain unit-volume. All models used $\gamma_0 = 0.01$.

The variance of each dimension $i$, $\sigma_{x_i}^2$, is guaranteed to be non-negative by using a softplus transfer function.

Whether trained with the na\"ive or intersection sampler, we evaluate model performance using
the na\"ive sampler with $200$ samples.  This approach ensures that we are comparing the quality of models
based only on the method by which they were trained.

\section{SPE Variants}
\label{appendix:spe_variants}
We assumed only diagonal covariance matrices in this work. Switching to a full
covariance matrix would require matrix inversion, which is ordinarily
infeasible, but because one purpose of deep embeddings is visualization, there
may be interesting cases involving 2D embeddings where the cost of inversion is
trivial. However, using a diagonal covariance matrix causes class-discriminating 
features to be aligned with the axes of the latent space, as we argued in the
main article, and this alignment is a virtue for interpretation.

\clearpage

\section{Tabular Results}
\label{appendix:tabular_results}

\subsection{Omniglot}

\begin{table}[h!]
\caption{Test classification accuracy (\%) on Omniglot with a $2$D embedding for clean-support/clean-query, corrupt-support/clean-query, and clean-support/corrupt-query. PN is our implementation of Prototypical Networks \citep{Snell2017}. SPE is our model. SPE is trained with intersection sampling (1 sample per trial). Reported accuracy for each experimental configuration is the mean over $1000$ random test episodes.}
\definecolor{hilitecolor3}{RGB}{235,235,235}
\newcommand{\hiz}[1]{\colorbox{hilitecolor3}{\parbox{\widthof{#1}}{#1}}}
\label{table:omni}
\begin{center}
\begin{small}
\begin{sc}
\begin{tabular}{cccccc}
\multicolumn{6}{l}{\hiz{Clean Support, Clean Query}} \\
\toprule
 & 1-shot, 5-class & 5-shot, 5-class & 1-shot, 20-class & 5-shot, 20-class & \hiz{Mean} \\
 \midrule
PN & 75.7 & 82.6 & 45.0 & 55.9 & \hiz{64.8} \\
SPE & 76.9 & 82.3 & 49.7 & 55.3 & \hiz{66.1} \\
\bottomrule
\end{tabular}
\vskip 0.1in
\begin{tabular}{cccccc}
\multicolumn{6}{l}{\hiz{Corrupt Support, Clean Query}} \\
\toprule
 & 1-shot, 5-class & 5-shot, 5-class & 1-shot, 20-class & 5-shot, 20-class & \hiz{Mean} \\
 \midrule
PN & 50.0 & 65.9 & 23.6 & 31.7 & \hiz{42.8} \\
SPE & 50.7 & 73.9 & 25.6 & 41.6 & \hiz{48.0} \\
\bottomrule
\end{tabular}
\vskip 0.1in
\begin{tabular}{cccccc}
\multicolumn{6}{l}{\hiz{Clean Support, Corrupt Query}} \\
\toprule
 & 1-shot, 5-class & 5-shot, 5-class & 1-shot, 20-class & 5-shot, 20-class & \hiz{Mean} \\
 \midrule
PN & 48.9 & 52.3 & 21.7 & 25.6 & \hiz{37.1} \\
SPE & 47.8 & 52.3 & 22.8 & 26.8 & \hiz{37.4} \\
\bottomrule
\end{tabular}
\end{sc}
\end{small}
\end{center}
\end{table}

\clearpage

\subsection{N-Digit MNIST}

\begin{table}[h!]
\caption{Test classification accuracy ($\%$) on $2$- and $3$-digit MNIST for 
clean-support/clean-query, corrupt-support/clean-query, and clean-support/corrupt-query. $N$: number of digits in each image; 
$D$: dimensionality of the embedding. Contrastive and HIB results from \cite{Oh2019}. PN is our implementation of Prototypical Networks \citep{Snell2017}. SPE is our model. SPE is trained with intersection sampling (1 sample per trial). Reported accuracy for PN and SPE for each experimental configuration is the mean over 1000 random test episodes.}
\label{table:2_digit_mnist}
\definecolor{hilitecolor3}{RGB}{235,235,235}
\newcommand{\hiz}[1]{\colorbox{hilitecolor3}{\parbox{\widthof{#1}}{#1}}}
\begin{center}
\begin{small}
\begin{sc}
\begin{tabular}{c|ccccc|ccccc}
\multicolumn{11}{l}{\hiz{Clean Support, Clean Query}} \\
\toprule
 & \multicolumn{5}{c|}{seen test classes} &  \multicolumn{5}{c}{unseen test classes}\\
\midrule
& \multicolumn{2}{c}{N=2} & \multicolumn{2}{c}{N=3} & & \multicolumn{2}{c}{N=2} & \multicolumn{2}{c}{N=3} \\
& D=2 & D=3 & D=2 & D=3 & \hiz{mean} & D=2 & D=3 & D=2 & D=3 & \hiz{mean}\\
\midrule
Contrastive & 88.2 & 95.0 & 65.8 & 87.3 & \hiz{84.1} & 85.5 & 84.8 & 59.0 & 85.5& \hiz{78.7} \\
HIB & 87.9 & 95.2 & 65.0 & 87.3 & \hiz{83.9} & 87.3 & 91.0 & 64.4 & 88.2 & \hiz{82.7}\\
PN & 91.1 & 95.0 & 65.8 & 90.6 & \hiz{85.6} & 82.0 & 89.5 & 64.3 & 89.1 & \hiz{81.2} \\
SPE & 93.0 & 94.2 & 80.2 & 89.0 & \hiz{89.1} & 90.0 & 89.3 & 80.2 & 88.2 & \hiz{86.9}\\
\bottomrule
\end{tabular}
\vskip 0.1in
\begin{tabular}{c|ccccc|ccccc}
\multicolumn{11}{l}{\hiz{Corrupt Support, Clean Query}} \\
\toprule
& \multicolumn{5}{c|}{seen test classes} &  \multicolumn{5}{c}{unseen test classes}\\
\midrule
& \multicolumn{2}{c}{N=2} & \multicolumn{2}{c}{N=3} & & \multicolumn{2}{c}{N=2} & \multicolumn{2}{c}{N=3} \\
& D=2 & D=3 & D=2 & D=3 & \hiz{mean} & D=2 & D=3 & D=2 & D=3 & \hiz{mean}\\
\midrule
Contrastive & 76.2 & 92.2 & 49.5 & 77.6 & \hiz{73.9} & 76.5 & 73.3 & 42.6 & 73.2 & \hiz{66.4} \\
HIB & 81.6 & 94.3 & 54.0 & 81.2 & \hiz{77.8} & 80.8 & 86.7 & 53.9 & 81.2 & \hiz{75.7}\\
PN & 72.7 & 93.3 & 44.6 & 82.7 & \hiz{73.3} & 70.9 & 86.3 & 42.9 & 79.6 & \hiz{69.9} \\
SPE & 92.4 & 93.8 & 76.7 & 87.8 & \hiz{87.7} & 88.8 & 86.3 & 75.4 & 86.3 & \hiz{84.2}\\
\bottomrule
\end{tabular}
\vskip 0.1in
\begin{tabular}{c|ccccc|ccccc}
\multicolumn{11}{l}{\hiz{Clean Support, Corrupt Query}} \\
\toprule
& \multicolumn{5}{c|}{seen test classes} &  \multicolumn{5}{c}{unseen test classes}\\
\midrule
& \multicolumn{2}{c}{N=2} & \multicolumn{2}{c}{N=3} & & \multicolumn{2}{c}{N=2} & \multicolumn{2}{c}{N=3} \\
& D=2 & D=3 & D=2 & D=3 & \hiz{mean} & D=2 & D=3 & D=2 & D=3 & \hiz{mean}\\
\midrule
Contrastive & 43.5 & 51.6 & 29.3 & 44.7 & \hiz{42.3} & 46.3 & 44.8 & 26.2 & 42.0 & \hiz{39.8} \\
HIB & 49.9 & 57.8 & 31.8 & 49.9 & \hiz{47.4} & 53.5 & 57.0 & 32.1 & 50.2 & \hiz{48.2}\\
PN & 53.1 & 61.1 & 33.8 & 56.4 & \hiz{51.1} & 51.1 & 57.9 & 33.0 & 54.8 & \hiz{49.2} \\
SPE & 53.7 & 58.2 & 40.2 & 48.1 & \hiz{50.1} & 56.3 & 56.5 & 39.3 & 46.6 & \hiz{49.7}\\
\bottomrule
\end{tabular}
\end{sc}
\end{small}
\end{center}
\end{table}

\clearpage

\section{Corruption Procedure}
\label{corruption_procedure}

\begin{SCfigure}[][h!]
    \centering
    \includegraphics[height=.4in]{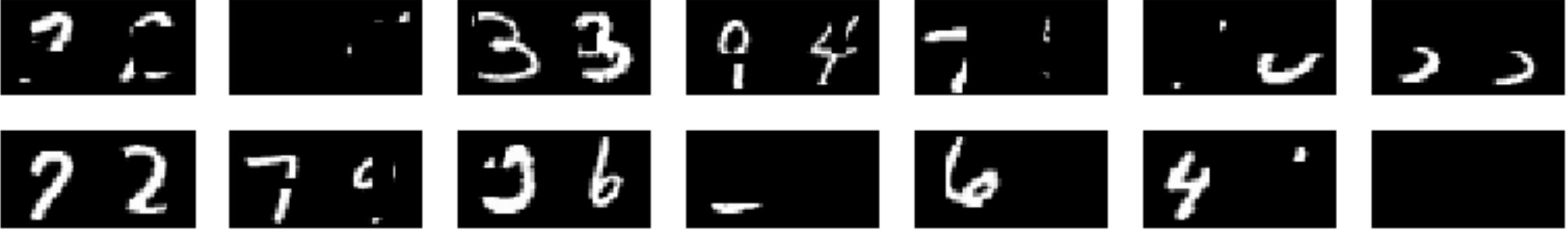}
    \caption{Examples of occluded 2-digit sequences. Occlusion is based on random rectangles
    that black out portions of each digit.}
    \label{fig:occluded}
\end{SCfigure}

The algorithm for applying corruption was identical to the scheme used in \cite{Oh2019}. A random rectangular-sized occlusion of black pixels was determined by first sampling a patch width, $L_x$, and patch height, $L_y$, from a uniform distribution, $L_x, L_y \sim \mathcal{U}(0, 28)$, and then sampling the top-left corner coordinates, $TL_x \sim \mathcal{U}(0, 28-L_x)$, $TL_y \sim \mathcal{U}(0, 28-L_y)$. This resulted in an occlusion of area $L_x \times L_y$. Note that if $L_x = 0$ or $L_y = 0$, the image was left unoccluded. Figure \ref{fig:occluded} shows examples of occluded 2-digit images. 

For Omniglot, we only trained/validated on corrupted imagery if the test set contained a corrupted support or corrupted query set. When testing on clean support and clean query, the training and validation sets were left unoccluded. When testing on corrupted imagery, the training and validation sets corrupted each character independently with a probability of $0.2$. 

The training and validation sets for $N$-digit MNIST corrupted each digit of each image independently with a probability of $0.2$, regardless of test imagery. This matched \cite{Oh2019}. 

During testing on both data sets, we considered both clean and corrupt support sets, as well as clean and corrupt query sets. A clean set was one in which all digits/characters were unoccluded. A corrupt set occluded each digit/character in each image according to the procedure described above.



\end{document}

%% file: math_commands.tex
\usepackage{amsmath,amsfonts,bm}

\def\eqref#1{equation~\ref{#1}}

\def\1{\bm{1}}

\def\vmu{{\bm{\mu}}}
\def\vsigma{{\bm{\sigma}}}

\def\vrho{{\bm{\rho}}}
\def\vepsilon{{\bm{\epsilon}}}

\def\vx{{\bm{x}}}

\def\vz{{\bm{z}}}

\def\mI{{\bm{I}}}

\DeclareMathAlphabet{\mathsfit}{\encodingdefault}{\sfdefault}{m}{sl}
\SetMathAlphabet{\mathsfit}{bold}{\encodingdefault}{\sfdefault}{bx}{n}

